\documentclass[runningheads]{llncs}

\usepackage{eccv}

\usepackage{eccvabbrv}

\usepackage{graphicx}
\usepackage{float}
\usepackage{booktabs}
\usepackage{multirow}
\usepackage[table]{xcolor}
\usepackage{makecell}
\usepackage{comment} 
\newif\ifincludesupp
\includesupptrue 
\usepackage[accsupp]{axessibility} 

\usepackage{hyperref}
\usepackage{adjustbox}
\usepackage{orcidlink}
\usepackage{bbding}

\begin{document}

\title{HiHR: Hierarchical Hyperbolic Representation for Aerial-Ground Person Re-Identification}

\titlerunning{HiHR: Hierarchical Hyperbolic Representation for AG-ReID}

\author{Qiwei Yang\orcidlink{0009-0007-3730-0903}, Pingping Zhang\orcidlink{0000-0003-1206-1444}\Envelope}

\authorrunning{Q.~Yang}

\institute{Dalian University of Technology, Dalian, China\\
  \Envelope\ Corresponding author: \email{zhpp@dlut.edu.cn}
}

\maketitle

\begin{abstract}
  Aerial-Ground Person Re-IDentification (AG-ReID) aims to retrieve the same person across heterogeneous aerial and ground camera platforms.
  Although great progress has been made, existing methods remain suboptimal due to the direct feature alignment across views, overlooking view-specific cues.
  To address this issue, we propose a novel Hierarchical Hyperbolic Representation (HiHR) framework for AG-ReID.
  More specifically, we first extract multi-granularity features based on pre-trained visual-text encoders.
  Then, we propose a Text-guided Multi-granularity Fusion (TMF) to fuse multi-granularity features and enhance the representation ability of identity features.
  Furthermore, we introduce the Hierarchical Hyperbolic Learning (HHL) to construct a hierarchical feature structure in a hyperbolic space.
  This hierarchy includes a coarse level that ensures identity separability and cross-view consistency, and a fine level that preserves view-specific discriminative cues.
  As a result, our proposed framework can effectively aggregate view-invariant and view-specific discriminative features for AG-ReID.
  Extensive experiments on four AG-ReID benchmarks demonstrate the effectiveness of our framework.
  The source code is available at \url{https://github.com/YangQiWei3/HiHR}.
\end{abstract}
\section{Introduction}
\label{sec:introduction}
\begin{figure}[tb]
  \centering
  \includegraphics[width=0.8\textwidth]{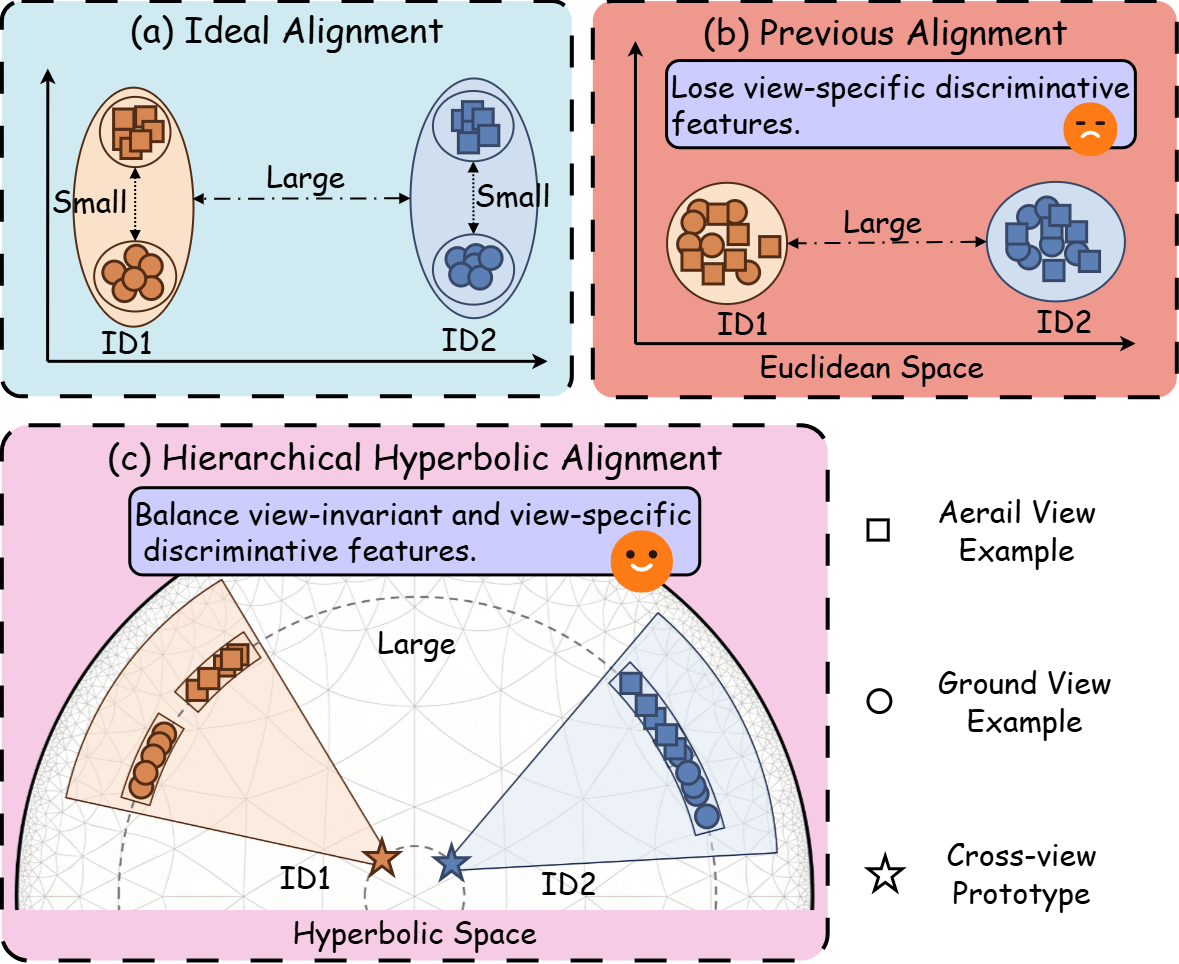}
  \caption{Illustration of our motivations and proposed framework.
    (a) The ideal alignment clusters examples of the same identity nearby while different identities far apart.
    (b) Previous alignments in Euclidean space align cross-view samples directly and may suppress view-specific discriminative cues.
    (c) Our hierarchical hyperbolic alignment keeps a hierarchical separability and view-specific discriminative cues.}
  \label{fig:motivation}
\end{figure}

Person Re-IDentification (ReID) aims to retrieve the same person across non-overlapping cameras, which serves as a fundamental capability for intelligent surveillance systems.
Despite great progress, existing methods~\cite{li2023clip,wang2025unity,wang2025makes} mainly focus on homogeneous camera networks, limiting their applicability to real-world scenarios.
In practice, real-world deployments increasingly integrate heterogeneous camera platforms, where ground cameras provide detailed close-range observations, while aerial cameras offer wide-area coverage.
This motivates Aerial-Ground Person Re-IDentification (AG-ReID), which aims to retrieve persons across heterogeneous platforms under systematic viewpoint discrepancies.
Such discrepancies introduce substantial appearance differences, hindering discriminative representation learning.
However, as shown in Fig.~\ref{fig:motivation}(a), the discriminative representation should be view-agnostic and allow cross-view discrepancies.
To this end, previous methods~\cite{wang2025secap,zhang2025latex,yang2026sas,wang2026sd,li2026gsalign} directly align samples of the same identity across views, as shown in Fig.~\ref{fig:motivation}(b).
Although effective, they emphasize view-invariant features while overlooking view-specific cues.
Moreover, many AG-ReID methods~\cite{zhang2024view,khalid2025bridging,huang2025perspective,ha2025multi,li2026gsalign} rely solely on global representations, which capture high-level semantics.
They may attenuate mid-level structural cues and fine-grained local patterns for reliable retrieval.

To address the aforementioned issues, we propose a novel Hierarchical Hyperbolic Representation (HiHR) framework for AG-ReID.
Specifically, inspired by previous works~\cite{li2023clip,wang2025makes}, we first introduce text prompts to extract multi-granularity features based on pre-trained visual-text encoders.
Then, we propose a Text-guided Multi-granularity Fusion (TMF) to fuse the multi-granularity features and enhance the representation ability of identity features.
Furthermore, we introduce the Hierarchical Hyperbolic Learning (HHL) to construct a hierarchical feature structure in a hyperbolic space. 
As illustrated in Fig.~\ref{fig:motivation}(c), it includes a coarse level for identity separability and cross-view consistency, together with a fine level for preserving view-specific discriminative cues.
In addition, the entailment regularization is further introduced to maintain the hierarchical consistency by propagating identity-consistent separation from the coarse level to the fine level.
As a result, our framework can effectively aggregate view-invariant and view-specific discriminative features for AG-ReID.
Extensive experiments on four AG-ReID benchmarks demonstrate the effectiveness of our framework.

In summary, our contributions are as follows:
\begin{itemize}
  \item We propose a novel Hierarchical Hyperbolic Representation (HiHR) framework for AG-ReID.
        To our best knowledge, it is the first work to utilize hyperbolic representations for cross-view person retrieval.
  \item We propose a Text-guided Multi-granularity Fusion (TMF) to fuse view-invariant and view-specific discriminative features in multi-granularity.
  \item We propose the Hierarchical Hyperbolic Learning (HHL) to model coarse-to-fine representation structures of cross-view and view-specific features.
  \item Extensive experiments on four benchmarks demonstrate that our proposed framework achieves competitive or state-of-the-art performance.
\end{itemize}
\section{Related Work}
\label{sec:related_work}
\subsection{Aerial-Ground Person Re-Identification}
AG-ReID extends conventional ground-to-ground person ReID~\cite{yan2023learning,wang2024other,wang2025decoupled,wang2025idea,huang2025generalizable} to heterogeneous aerial-ground camera networks.
Early works~\cite{nguyen2023aerial,nguyen2024ag,zhang2024view,wang2025secap} mainly aim to establish benchmarks and baselines.
Building on these foundations, recent methods~\cite{zhang2024view,zhu2025global,huang2025perspective,li2026gsalign,mao2026cvaf,wang2026sd,zhang2026view} propose view-aware models to mitigate the aerial-ground viewpoint discrepancies.
Technically, Zhang \etal~\cite{zhang2024view} propose a view-decoupled Transformer to separate view-related and view-unrelated information.
Huang \etal~\cite{huang2025perspective} propose a perspective-driven prototype alignment to stabilize cross-view matching.
Li \etal~\cite{li2026gsalign} develop a geometric-and-semantic alignment network to compensate cross-view distortion and emphasize visibility-consistent regions.
Mao \etal~\cite{mao2026cvaf} introduce learnable text tokens to enhance cross-view feature consistency.
Zhang \etal~\cite{zhang2025latex} adopt prompt-tuning strategies to leverage attribute-based text knowledge for robust representations. 
Wang \etal~\cite{wang2026sd} leverage stable diffusion models to mimic the feature distribution of different views while extracting robust identity representations.
Zhang \etal~\cite{zhang2026view} propose a view-aware semantic alignment framework to achieve cross-view semantic consistency.
Despite promising results, they directly align samples of the same identity across views, overlooking view-specific discriminative information.
This limitation motivates our framework, which aims to maintain cross-view identity consistency while preserving view-specific discriminative cues.
\subsection{Hyperbolic Representation Learning for ReID}
Hyperbolic representation learning provides an effective inductive bias for hierarchical or tree-structured data, by leveraging the exponential volume growth induced by negative curvatures~\cite{cannon1997hyperbolic,lee2006riemannian}.
This property has motivated a growing body of work that leverages hyperbolic geometry to better preserve semantic granularity, improve metric learning and enhance representation separability.
However, in person ReID, hyperbolic representation learning has not been fully explored.
Early attempts mainly adopt hyperbolic geometry as a drop-in replacement of the Euclidean embedding space.
For example, Khrulkov \etal~\cite{khrulkov2020hyperbolic} append hyperbolic layers to standard visual encoders and project image features into a Poincar\'e ball, where the similarity is computed with a hyperbolic distance.
Fang \etal~\cite{fang2021kernel} develop positive definite kernels for hyperbolic spaces and evaluate them on person ReID tasks.
Although effective, existing methods usually treat the hyperbolic representation as a single-level embedding or similarity metric.
This makes them difficult to capture persons' complex appearance variations caused by viewpoint discrepancies.
To address this limitation, we propose HHL to embed multi-granularity representations into a Lorentz hyperbolic manifold~\cite{cannon1997hyperbolic}, thereby modeling the coarse-to-fine feature hierarchy for AG-ReID.
\begin{figure}[t]
  \centering
  \includegraphics[width=1.0\textwidth]{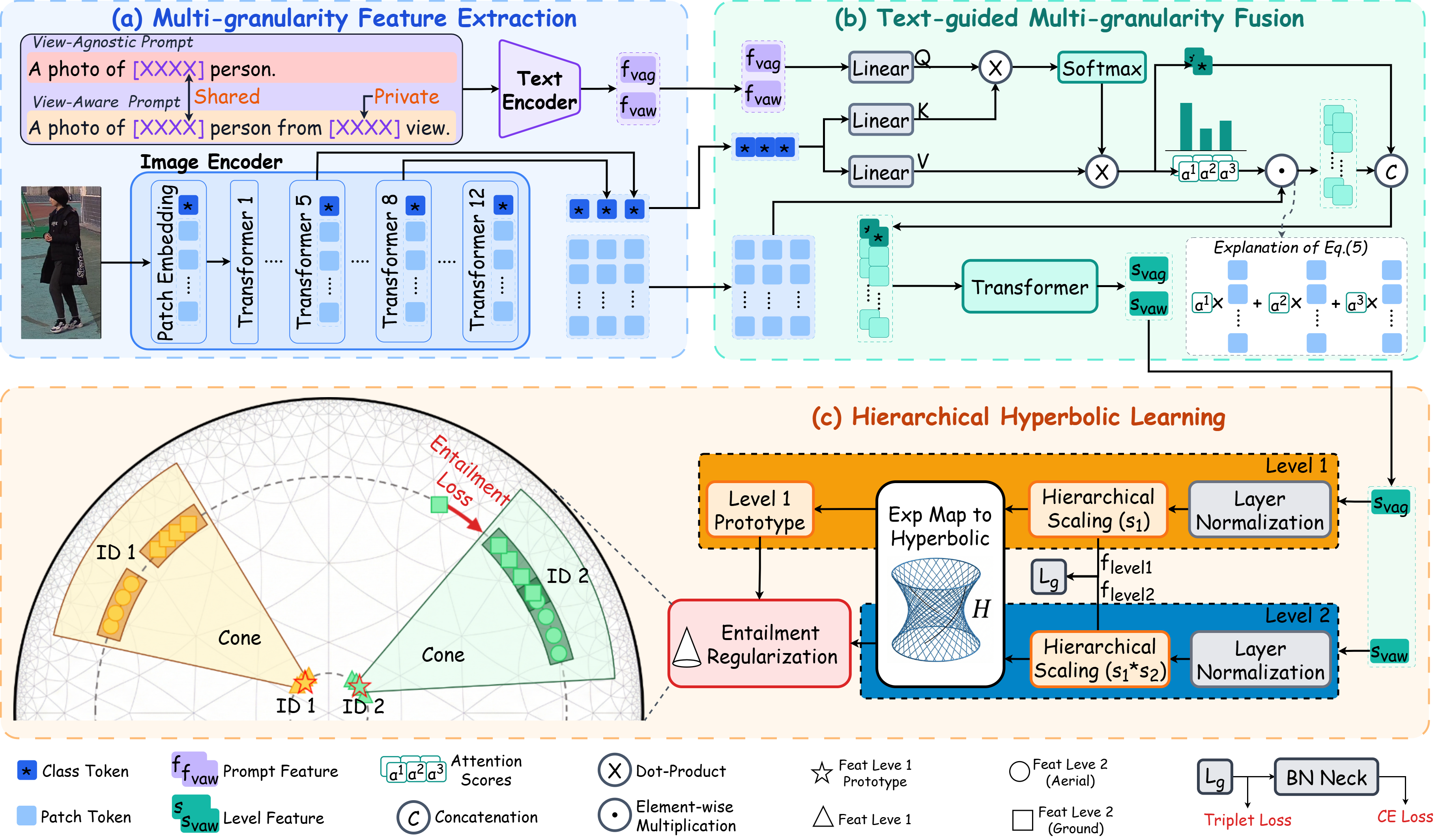}
  \caption{Illustration of our proposed framework.
    It includes three key modules, i.e., Multi-granularity Feature Extraction (MFE), Text-guided Multi-granularity Fusion (TMF) and Hierarchical Hyperbolic Learning (HHL).
  }
  \label{fig:overall}
  \vspace{-4mm}
\end{figure}
\section{Methodology}
\label{sec:method}
In this section, we introduce the proposed Hierarchical Hyperbolic Representation (HiHR) framework for AG-ReID.
As shown in Fig.~\ref{fig:overall}, it includes three key modules, i.e., Multi-granularity Feature Extraction (MFE), Text-guided Multi-granularity Fusion (TMF) and Hierarchical Hyperbolic Learning (HHL).
Details are described as follows.
\subsection{Multi-granularity Feature Extraction}
In practice, many AG-ReID methods rely solely on global representations, which capture high-level semantics.
However, they may attenuate mid-level structural cues and fine-grained local patterns for reliable retrieval.
To improve the feature representation ability, we first introduce the Multi-granularity Feature Extraction (MFE).
More specifically, inspired by previous works~\cite{li2023clip,wang2025makes}, we extract multi-granularity features based on pre-trained visual-text encoders, such as CLIP~\cite{radford2021learning}.
For visual features, we utilize the output of different Transformer layers from the image encoder $E_v$.
For an input image $\mathbf{I}$, the token sequence at the $l$-th Transformer layer is denoted as
\begin{equation}
  \mathbf{Z}^{(l)}=[\mathbf{c}^{(l)};\mathbf{X}^{(l)}]\in\mathbb{R}^{(1+N)\times D},
\end{equation}
where $\mathbf{c}^{(l)}$ and $\mathbf{X}^{(l)}$ denote the class token and patch tokens, respectively.
$N$ is the number of patch tokens.
$D$ is the token dimension.

For text features, we introduce dual-level prompts to extract view-agnostic and view-aware information.
As shown in Fig.~\ref{fig:overall}(a), the view-agnostic prompt $\mathbf{p}_{vag}$ is defined as ``\textit{A photo of [\emph{shared}] person.}''.
It emphasizes identity-consistent person evidence.
The corresponding embedding is obtained by the text encoder $E_t$ as follows:
\begin{equation}
  \mathbf{f}_{vag}=E_t(\mathbf{p}_{vag})\in\mathbb{R}^{D}.
\end{equation}
In parallel, the view-aware prompt $\mathbf{p}_{vaw}$ is defined as ``\textit{A photo of [\emph{shared}] person from [\emph{view-private}] view.}''.
It explicitly contains viewpoint information and captures view-specific discriminative cues within the corresponding view.
The corresponding embedding is obtained as follows:
\begin{equation}
  \mathbf{f}_{vaw}=E_t(\mathbf{p}_{vaw})\in\mathbb{R}^{D}.
\end{equation}
Thus, MFE yields multi-granularity visual features $\mathbf{Z}=[\mathbf{Z}^{(l_1)};\mathbf{Z}^{(l_2)};\dots;\mathbf{Z}^{(l_M)}]$ and text features $\{\mathbf{f}_{vag}, \mathbf{f}_{vaw}\}$.
Here, $M$ is the total number of selected layers.
They are complementary inputs to the subsequent modules.
\subsection{Text-guided Multi-granularity Fusion}
Technically, one can concatenate the above multi-granularity features, and perform person retrieval.
However, there are modality gaps in the visual features and text features, which may lead to inferior performance.
To address this issue, we propose a Text-guided Multi-granularity Fusion (TMF) to fuse multi-granularity features and enhance the representation ability of identity features.

To integrate complementary information from different granularities, we perform a cross-attention by using the text features as semantic queries and the class tokens from different Transformer layers as keys and values, as follows:
\begin{equation}
  \boldsymbol{\alpha}_{k}=\mathrm{softmax}\!\left(\frac{\mathbf{Q}_{k}\mathbf{K}^{\top}}{\sqrt{D}}\right),\quad
  \mathbf{t}_{k}=\boldsymbol{\alpha}_{k}\mathbf{V},\quad k\in\{vag,vaw\},
  \label{eq:cls_fusion}
\end{equation}
where $(\mathbf{Q}_{k},\mathbf{K},\mathbf{V})$ are linear projections of $(\mathbf{f}_{k},\mathbf{C},\mathbf{C})$.
$\mathbf{C}=[\mathbf{c}^{(l_1)};\mathbf{c}^{(l_2)};\dots;\mathbf{c}^{(l_M)}]$.
Thus, we can obtain the view-agnostic fused feature $\mathbf{t}_{vag}$ and the view-aware fused feature $\mathbf{t}_{vaw}$, respectively.

Meanwhile, fine-grained discriminative cues often reside in patch tokens~\cite{zhang2024magic}.
Thus, we reuse $\boldsymbol{\alpha}_{k}$ to aggregate multi-layer patch tokens as follows:
\begin{equation}
  \mathbf{X}_{k}=\sum_{m=1}^{M}\alpha^{(m)}_{k}\,\mathbf{X}^{(l_m)}.
  \label{eq:patch_fusion}
\end{equation}
We then fuse $\mathbf{X}_{k}$ with $\mathbf{t}_{k}$ through a standard Transformer layer:
\begin{equation}
  {\mathbf{S}}_{k}=\mathrm{Trans}\big([\mathbf{t}_{k};\mathbf{X}_{k}]\big).
  \label{eq:tmf_output}
\end{equation}
We take the first token ${\mathbf{s}}_{k}\in\mathbb{R}^{D}$ in ${\mathbf{S}}_{k}$ as the output representation.
As observed, our TMF can adaptively fuse multi-granularity features.
The fused features provide a better representation for robust cross-view retrieval and serve as the inputs to the subsequent hierarchical hyperbolic learning.

\subsection{Hierarchical Hyperbolic Learning}
In fact, AG-ReID needs to maintain cross-view identity consistency and view-specific discriminative cues.
To this end, we propose the Hierarchical Hyperbolic Learning (HHL) to construct a hierarchical feature structure in a hyperbolic space.
As shown in Fig.~\ref{fig:motivation}(c), it includes a coarse level that ensures identity separability and cross-view consistency, and a fine level that preserves view-specific discriminative cues.
In addition, we introduce the entailment regularization to ensure different hierarchical structures.

\textbf{Hyperbolic Representation.}
A hyperbolic space is a smooth Riemannian manifold with a constant negative sectional curvature $-\tau$ ($\tau>0$)~\cite{lee2006riemannian, cannon1997hyperbolic}.
To obtain the hyperbolic representation, we define the following formulas.
The exponential map $\mathrm{exp}^{\tau}_{\mathbf{x}}(\cdot)$ projects tangent vectors at $\mathbf{x}$ to a Lorentz manifold~\cite{cannon1997hyperbolic}.
The logarithmic map $\mathrm{log}^{\tau}_{\mathbf{y}}(\cdot)$ projects points on the Lorentz space to the tangent space at $\mathbf{y}$.
For each $\mathbf{x}$, the hyperbolic entailment cone $\omega(\mathbf{x})$ specify a region whose points $\mathbf{y}\in\omega(\mathbf{x})$ are treated as child embeddings of $\mathbf{x}$.
The cone is defined by half-aperture
\begin{equation}
  \omega(\mathbf{x})
  =\sin^{-1}\!\left(\frac{2k}{\sqrt{\tau}\,\|\mathbf{x}_{\mathrm{space}}\|}\right),
  \label{eq:cone}
\end{equation}
where $k=0.1$ is a constant and $\mathbf{x}_{\mathrm{space}}$ denotes the spatial coordinates of $\mathbf{x}$.
Hyperbolic entailment cones specify a region whose aperture is denoted by $\omega(\mathbf{x})$.
To ensure the child embeddings $\mathbf{y}$ within the parent cones $\omega(\mathbf{x})$, one can define the entailment loss as
\begin{equation}
  \mathcal{L}_{\mathrm{e}}(\mathbf{x},\mathbf{y})
  =\max\big(0,\phi(\mathbf{x},\mathbf{y})-\omega(\mathbf{x})\big),
  \label{eq:entail_loss}
\end{equation}
where $\phi(\mathbf{x},\mathbf{y})$ denotes the angular deviation of $\mathbf{y}$ from the axis of the entailment cone centered at $\mathbf{x}$.

For HHL, we take the two complementary embeddings from TMF as the parent and child representations.
To reflect the coarse-to-fine hierarchy, we normalize the two representations and impose learnable scale separations:
\begin{equation}
  \mathbf{g}^{p}=s_1\frac{{\mathbf{s}}_{vag}}{\|{\mathbf{s}}_{vag}\|_2},\qquad
  \mathbf{g}^{c}=s_1s_2\frac{{\mathbf{s}}_{vaw}}{\|{\mathbf{s}}_{vaw}\|_2},\qquad
  s_1>0,s_2>1,
\end{equation}
where $s_1$ controls the global embedding scale and $s_2$ pushes child embeddings farther from the origin than parent embeddings.
This scale separation provides a stable hierarchical initialization and reduces interference between the two levels.
\begin{figure}[tb]
  \centering
  \includegraphics[width=0.7\columnwidth]{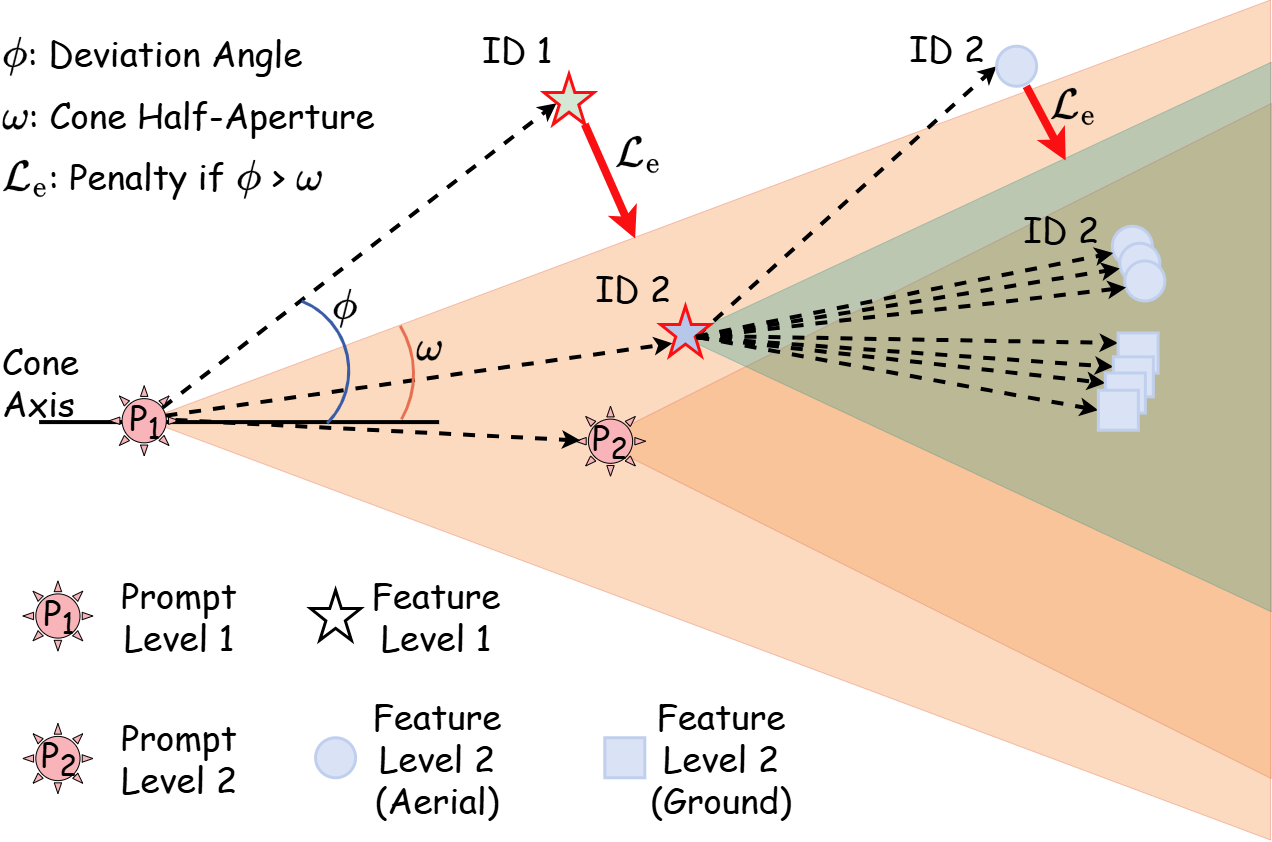}
  \caption{Illustration of entailment regularizations for hierarchical consistency in HHL.
  Four entailment constraints are imposed: parent prototypes to child embeddings ($\mathbf{z}^{\boldsymbol{\pi}}\!\rightarrow\!\mathbf{z}^{c}$), parent prompts to child prompts ($\mathbf{z}^{p}_{P}\!\rightarrow\!\mathbf{z}^{c}_{P}$), parent prompts to parent visual features ($\mathbf{z}^{p}_{P}\!\rightarrow\!\mathbf{z}^{p}$), and child prompts to child visual features ($\mathbf{z}^{c}_{P}\!\rightarrow\!\mathbf{z}^{c}$).
  These constraints jointly preserve hierarchical consistency, align vision-text semantics, and limit the drift of child embeddings while retaining view-specific discriminative cues.}
  \label{fig:loss}
\end{figure}

\textbf{Entailment Regularization for Hierarchical Structure.}
The above scale separation provides an initial coarse-to-fine hierarchy, but child embeddings may still drift and overlap across identities.
To impose identity-aware hierarchical constraints, we first compute parent prototypes from the scaled parent features in Euclidean space.
For each identity $y$ in a batch, the parent prototype $\boldsymbol{\pi}^{p}_{y}$ is defined as
\begin{equation}
  \boldsymbol{\pi}^{p}_{y}=\mathrm{Mean}\big(\{\mathbf{g}^{p}\mid y\}\big),
\end{equation}
where $\mathrm{Mean}(\cdot)$ denotes the mean operation.
Each prototype serves as the center of an identity-specific parent cone.
We then project $\mathbf{g}^{p}$, $\mathbf{g}^{c}$ and $\boldsymbol{\pi}^{p}_{y}$ to the Lorentz hyperbolic space using the exponential map $\mathrm{exp}^{\tau}_{\mathbf{x}}(\cdot)$ at the origin $\mathbf{o}=[\frac{1}{\sqrt{\tau}},\mathbf{0}]$:
\begin{equation}
  \mathbf{z}^{p}=\mathrm{exp}^{\tau}_{\mathbf{o}}(\mathbf{g}^{p}),\qquad
  \mathbf{z}^{c}=\mathrm{exp}^{\tau}_{\mathbf{o}}(\mathbf{g}^{c}),\qquad
  \mathbf{z}^{\boldsymbol{\pi}}=\mathrm{exp}^{\tau}_{\mathbf{o}}(\boldsymbol{\pi}^{p}_{y}).
\end{equation}
Given the prototype $\mathbf{z}^{\boldsymbol{\pi}}$, its entailment cone is centered at $\mathbf{z}^{\boldsymbol{\pi}}$, with the aperture $\omega(\mathbf{z}^{\boldsymbol{\pi}})$ determined by the prototype distance to the origin.
The child embedding $\mathbf{z}^{c}$ is regularized to lie within the corresponding identity-specific cone:
\begin{equation}
  \mathcal{L}_{\mathrm{e}}(\mathbf{z}^{\boldsymbol{\pi}},\mathbf{z}^{c})
  =\max\big(0,\phi(\mathbf{z}^{\boldsymbol{\pi}},\mathbf{z}^{c})-\omega(\mathbf{z}^{\boldsymbol{\pi}})\big).
\end{equation}
This constraint confines child-level refinement to the parent-induced entailment cone, preserving view-specific discriminative cues while mitigating uncontrolled drift and cross-identity confusion.

\textbf{Entailment Regularization for Visual-Prompt Consistency.}
Beyond the visual prototype-to-child constraint, we further align the prompt hierarchy with the visual hierarchy.
The view-agnostic feature $\mathbf{f}_{vag}$ provides broader semantic guidance than the view-aware feature $\mathbf{f}_{vaw}$.
This is because $\mathbf{f}_{vag}$ encodes general identity-related semantics shared across views, whereas $\mathbf{f}_{vaw}$ incorporates view-conditioned semantic details.
Accordingly, we treat $\mathbf{f}_{vag}$ as the parent prompt and regularize it to entail the child prompt $\mathbf{f}_{vaw}$.
Following the previous scale separation, the two features are projected to the hyperbolic space as
\begin{equation}
  \mathbf{z}^{p}_{P}=\mathrm{exp}^{\tau}_{\mathbf{o}}(s_1\frac{\mathbf{f}_{vag}}{\|\mathbf{f}_{vag}\|_2}),\qquad
  \mathbf{z}^{c}_{P}=\mathrm{exp}^{\tau}_{\mathbf{o}}(s_1s_2\frac{\mathbf{f}_{vaw}}{\|\mathbf{f}_{vaw}\|_2}),
\end{equation}
and the prompt-level hierarchy is enforced by $\mathcal{L}_{\mathrm{e}}(\mathbf{z}^{p}_{P},\mathbf{z}^{c}_{P})$.

Since text generally provides broader context than images~\cite{pal2025compositional}, we further constrain visual embeddings to be entailed by their same-level prompt embeddings.
This yields the parent-level and child-level visual-semantic constraints $\mathcal{L}_{\mathrm{e}}(\mathbf{z}^{p}_{P},\mathbf{z}^{p})$ and $\mathcal{L}_{\mathrm{e}}(\mathbf{z}^{c}_{P},\mathbf{z}^{c})$, respectively.
Together with the prototype-to-child constraint, these four entailment constraints form the hierarchical structure, as shown in Fig.~\ref{fig:loss}.
They preserve the semantic consistency between prompts and visual representations.

\textbf{Hierarchical Supervision.}
In addition to the entailment regularizations, we employ standard ReID objectives to supervise the two hierarchy levels.
Given an input feature $\mathcal{F}$, the objective combines the label-smoothing cross-entropy loss $\mathcal{L}_{\mathrm{CE}}$~\cite{szegedy2016rethinking} and triplet loss $\mathcal{L}_{\mathrm{Tri}}$~\cite{hermansdefense}:
\begin{equation}
  \mathcal{L}_{g}(\mathcal{F})=\lambda_g\mathcal{L}_{\mathrm{CE}}(\mathcal{F})+\mathcal{L}_{\mathrm{Tri}}(\mathcal{F}),
  \label{eq:base_loss}
\end{equation}
where $\lambda_g$ balances the cross-entropy and triplet losses.
At the parent level, $\mathcal{L}_{g}(\mathbf{g}^{p})$ is applied with cross-view mixed sampling to learn view-invariant identity semantics.
This supervision compacts same-identity samples across aerial and ground views, thereby strengthening coarse-level cross-view consistency.
At the child level, $\mathcal{L}_{g,\mathrm{intra}}(\mathbf{g}^{c})$ is applied within each view to avoid over-suppressing view-dependent variations:
\begin{equation}
  \mathcal{L}_{g,\mathrm{intra}}(\mathbf{g}^{c})=\lambda_g\mathcal{L}_{\mathrm{CE}}(\mathbf{g}^{c})+\mathcal{L}_{\mathrm{Tri},\mathrm{intra}}(\mathbf{g}^{c}).
\end{equation}
Compare with standard $\mathcal{L}_{\mathrm{Tri}}$, positives and negatives in $\mathcal{L}_{\mathrm{Tri,intra}}(\mathbf{g}^{c})$ are sampled from the same view as the anchor.
This intra-view supervision preserves view-specific discriminative cues and refines the same-view identity structure.
Together, the parent and child objectives impose cross-view compactness at the coarse level and view-aware compactness at the fine level.
\subsection{Training and Inference}
\textbf{Overall Objective.}
The final training objective is a weighted combination of the above losses and entailment regularizations:
\begin{equation}
  \begin{aligned}
    \mathcal{L}
     & =\mathcal{L}_{g}(\mathbf{g}^{p})
    +\mathcal{L}_{g,\mathrm{intra}}(\mathbf{g}^{c}) \\
     & \quad+\lambda_e\Big(
    \mathcal{L}_{\mathrm{e}}(\mathbf{z}^{\boldsymbol{\pi}},\mathbf{z}^{c})
    +\mathcal{L}_{\mathrm{e}}(\mathbf{z}^{p}_{P},\mathbf{z}^{c}_{P})
    +\mathcal{L}_{\mathrm{e}}(\mathbf{z}^{p}_{P},\mathbf{z}^{p})
    +\mathcal{L}_{\mathrm{e}}(\mathbf{z}^{c}_{P},\mathbf{z}^{c})
    \Big),
  \end{aligned}
  \label{eq:full_obj}
\end{equation}
where $\lambda_e$ is a hyperparameter that balances the objectives.

\textbf{Inference.}
During testing, we map $\mathbf{z}^{p}$ and $\mathbf{z}^{c}$ back to the Euclidean space using the logarithmic map $\mathrm{log}^{\tau}_{\mathbf{o}}(\cdot)$.
Then, we reduce them by ${s_1}$ and ${s_1s_2}$, respectively, and concatenate them to obtain the final retrieval feature.
\section{Experiments}
In this section, we first introduce the datasets, evaluation protocols and implementation details.
Then, we compare our method with state-of-the-arts.
Finally, ablation studies are conducted to verify the contribution of each component.
\subsection{Datasets and Evaluation Protocols.}
We evaluate our framework on four AG-ReID benchmarks.
To be specific, the AG-ReID v1~\cite{nguyen2023ag} includes 21,983 images of 388 identities captured by one ground camera and one aerial camera.
The AG-ReID v2~\cite{nguyen2024ag} comprises 100,502 images of 1,615 identities collected from three platforms, including UAV, wearable cameras and CCTV.
The LAGPeR~\cite{wang2025secap} contains 63,841 images of 4,231 identities collected by 14 ground cameras and seven aerial cameras.
The CARGO~\cite{zhang2024view} is a large-scale synthetic benchmark with 108,563 images of 5,000 identities from eight ground cameras and five aerial cameras.
For each benchmark, we strictly follow the official data splits and evaluation protocols for fair comparison.
In this context, \textbf{G} stands for ground, \textbf{A} for aerial, \textbf{C} for CCTV, \textbf{W} for wearable and \textbf{ALL} for view-agnostic protocols.
Following previous works~\cite{zhang2024view,wang2026sd}, the performance is evaluated by using Cumulative Matching Characteristics (CMC) at Rank-1 and mean Average Precision (mAP).
\subsection{Implementation Details.}
Our framework is implemented in PyTorch and trained on an NVIDIA A800 GPU.
We adopt the CLIP-ViT-B/16~\cite{radford2021learning} as the visual and text encoder.
For all datasets, images are resized to $256 \times 128 \times 3$.
The data augmentation includes random horizontal flipping, random cropping and random erasing.
We train for 60 epochs using Adam~\cite{kingma2014adam} with a batch size of 64, with 4 images sampled per identity.
For the AG-ReID v2 dataset, we increase the batch size to 128.
The learning rate is set to $3.5 \times 10^{-4}$ for randomly initialized modules and $5 \times 10^{-6}$ for pre-trained components.
A warm-up strategy with 100 iterations is applied before the cosine learning rate schedule.
Experimentally, we extract multi-granularity visual tokens from layers $\{5,8,12\}$ of the CLIP image encoder, and set $\lambda_e=5$, and $\lambda_g=1.0/0.25$ for $\mathcal{L}_{g}$/$\mathcal{L}_{g,\mathrm{intra}}$ as default.
In addition, we initialize $(s_1,s_2)=(0.5,2.0)$ and $\tau=1.0$.
The scaling factors and curvature are optimized end-to-end during training.
\subsection{Comparison with State-of-the-Arts}
As shown in Tab.~\ref{tab:sota_agreidv1_agreidv2}-Tab.~\ref{tab:sota_cargo}, we compare our method with state-of-the-arts on four AG-ReID benchmarks.
Our method achieves consistent advantages across all benchmarks.
On AG-ReID v1 and AG-ReID v2, it obtains the best mAP under all six evaluation settings.
On LAGPeR, our method achieves the best results across all three settings, further validating its effectiveness under diverse aerial-ground view compositions.
On CARGO, our method also ranks first in mAP across three protocols.
It verifies the generalization on synthetic datasets.
Notably, under the challenging A$\rightarrow$G setting, it reaches 70.53 mAP and 75.53 Rank-1, outperforming the strongest baselines by +1.53 mAP and +4.25 Rank-1.
The most notable advantage appears in mAP, indicating that our method improves the overall ranking quality rather than only the top-1 match.
This gain is consistent with our hierarchical alignment strategy, which fuses view-invariant and view-specific discriminative features.
By avoiding feature collapse with cross-view supervision, our method maintains more reliable similarity ordering with background interference, scale variation, and severe viewpoint changes.
Such full-list retrieval improvements suggest strong potential for practical aerial-ground search, where multiple correct candidates beyond the first rank are valuable.
\begin{table}[t]
  \caption{\textbf{Performance comparison on AG-ReID v1 and AG-ReID v2.}
    Best results are in \textbf{bold}.}
  \label{tab:sota_agreidv1_agreidv2}
  \centering
  \begin{adjustbox}{max width=\textwidth}
    \begin{tabular}{lcccccccccccc}
      \Xhline{1.0pt}
      \multirow{4}{*}{Method}
                                      & \multicolumn{4}{c}{AG-ReID v1}
                                      & \multicolumn{8}{c}{AG-ReID v2}                                                                              \\
      \cmidrule(lr){2-5}\cmidrule(lr){6-13}
                                      & \multicolumn{2}{c}{A$\rightarrow$G}
                                      & \multicolumn{2}{c}{G$\rightarrow$A}
                                      & \multicolumn{2}{c}{A$\rightarrow$C}
                                      & \multicolumn{2}{c}{C$\rightarrow$A}
                                      & \multicolumn{2}{c}{A$\rightarrow$W}
                                      & \multicolumn{2}{c}{W$\rightarrow$A}                                                                                       \\
      \cmidrule(lr){2-3}\cmidrule(lr){4-5}\cmidrule(lr){6-7}\cmidrule(lr){8-9}\cmidrule(lr){10-11}\cmidrule(lr){12-13}
                                      & \multicolumn{1}{c}{mAP}             & \multicolumn{1}{c}{R1} & \multicolumn{1}{c}{mAP} & \multicolumn{1}{c}{R1} & \multicolumn{1}{c}{mAP} & \multicolumn{1}{c}{R1} & \multicolumn{1}{c}{mAP} & \multicolumn{1}{c}{R1} & \multicolumn{1}{c}{mAP} & \multicolumn{1}{c}{R1} & \multicolumn{1}{c}{mAP} & \multicolumn{1}{c}{R1} \\
      \Xhline{0.6pt}
      Explain~\cite{nguyen2023aerial} & 72.38                               & 81.28                  & 73.35                   & 82.64                  & 79.00                   & 87.70                  & 78.24                   & 87.35                  & 83.14                   & \textbf{93.67}         & 79.08                   & 87.73                  \\
      VDT~\cite{zhang2024view}        & 74.44                               & 82.91                  & 78.57                   & 86.59                  & 79.13                   & 86.46                  & 78.12                   & 86.14                  & 82.21                   & 90.00                  & 78.52                   & 85.26                  \\
      GSAlign~\cite{li2026gsalign}    & 75.01                               & 83.75                  & 77.73                   & 84.10                  & 81.38                   & 87.86                  & 81.05                   & 88.02                  & 83.98                   & 90.63                  & 80.90                   & 87.31                  \\
      SD-ReID~\cite{wang2026sd}       & 75.40                               & 85.16                  & 77.02                   & 85.57                  & 80.61                   & 87.04                  & 79.24                   & 86.74                  & 84.06                   & 90.86                  & 80.12                   & 86.79                  \\
    ViSA~\cite{zhang2026view}       & 75.56                               & 84.91                  & 77.94                   & 84.10                  & 83.61                   & \textbf{89.43}         & 82.32                   & 88.57                  & 85.99                   & 91.63                  & 83.10                   & 89.23                  \\
      SAS-VPReID~\cite{yang2026sas}   & 75.98                               & 84.24                  & 77.28                   & 86.17                  & 76.39                   & 83.40                  & 78.47                   & 85.26                  & 82.10                   & 89.45                  & 78.47                   & 85.73                  \\
      SeCap~\cite{wang2025secap}      & 76.16                               & 84.03                  & 78.34                   & 87.01                  & 80.84                   & 88.12                  & 79.99                   & 88.24                  & 84.01                   & 91.44                  & 80.15                   & 87.56                  \\
      LATex~\cite{zhang2025latex}     & 77.67                               & 85.26                  & 81.15                   & \textbf{89.40}         & 83.50                   & 89.13         & 82.85                   & \textbf{89.01}         & 86.35                   & 91.35                  & 83.30                   & 89.32                  \\
      \Xhline{0.6pt}
      \rowcolor{gray!15}
      \textbf{HiHR}                   & \textbf{79.21}                      & \textbf{86.06}         & \textbf{81.28}          & 87.94                  & \textbf{84.04}          & 88.84                  & \textbf{83.21}          & 88.57                  & \textbf{87.57}          & 91.40                  & \textbf{84.02}          & \textbf{89.53}         \\
      \Xhline{1.0pt}
    \end{tabular}
  \end{adjustbox}
  \vspace{-2mm}
\end{table}
\begin{table}[t]
  \caption{\textbf{Performance comparison on LAGPeR.}
    Best results are in \textbf{bold}.}
  \label{tab:sota_lagper}
  \centering
  \begin{tabular}{lcccccc}
    \Xhline{1.0pt}
    \multirow{2}{*}{Method}
                                    & \multicolumn{2}{c}{A$\rightarrow$G}
                                    & \multicolumn{2}{c}{G$\rightarrow$A}
                                    & \multicolumn{2}{c}{G$\rightarrow$AG}                                                                                                                                \\
    \cmidrule(lr){2-3}\cmidrule(lr){4-5}\cmidrule(lr){6-7}
                                    & \multicolumn{1}{c}{mAP}              & \multicolumn{1}{c}{R1} & \multicolumn{1}{c}{mAP} & \multicolumn{1}{c}{R1} & \multicolumn{1}{c}{mAP} & \multicolumn{1}{c}{R1} \\
    \Xhline{0.6pt}
    Explain~\cite{nguyen2023aerial} & 28.89                                & 40.48                  & 31.91                   & 32.96                  & 17.89                   & 22.03                  \\
    VDT~\cite{zhang2024view}        & 28.97                                & 40.15                  & 31.98                   & 33.55                  & 16.45                   & 19.50                  \\
    SD-ReID~\cite{wang2026sd}       & 29.72                                & 40.15                  & 32.86                   & 34.47                  & 18.88                   & 22.85                  \\
    SeCap~\cite{wang2025secap}      & 30.37                                & 41.79                  & 33.42                   & 35.26                  & 19.24                   & 24.39                  \\
    ViSA~\cite{zhang2026view}       & 30.42                                & 41.37                  & 33.59                   & 35.06                  & 20.31                   & 25.61                  \\
    SAS-VPReID~\cite{yang2026sas}   & 30.76                                & 43.83                  & 33.40                   & 35.42                  & 23.02                   & 31.12                  \\
    \Xhline{0.6pt}
    \rowcolor{gray!15}
    \textbf{HiHR}                   & \textbf{33.80}                       & \textbf{46.95}         & \textbf{36.74}          & \textbf{39.30}         & \textbf{23.99}          & \textbf{31.39}         \\
    \Xhline{1.0pt}
  \end{tabular}
  \vspace{-2mm}
\end{table}
\begin{table}[t]
  \caption{\textbf{Performance comparison on CARGO.}
    Best results are in \textbf{bold}.}
  \label{tab:sota_cargo}
  \centering
  \begin{tabular}{lcccccccc}
    \Xhline{1.0pt}
    \multirow{2}{*}{Method}
                                  & \multicolumn{2}{c}{ALL}
                                  & \multicolumn{2}{c}{A$\rightarrow$G}
                                  & \multicolumn{2}{c}{A$\rightarrow$A}
                                  & \multicolumn{2}{c}{G$\rightarrow$G}                                                                                                                                                                                   \\
    \cmidrule(lr){2-3}\cmidrule(lr){4-5}\cmidrule(lr){6-7}\cmidrule(lr){8-9}
                                  & \multicolumn{1}{c}{mAP}             & \multicolumn{1}{c}{R1} & \multicolumn{1}{c}{mAP} & \multicolumn{1}{c}{R1} & \multicolumn{1}{c}{mAP} & \multicolumn{1}{c}{R1} & \multicolumn{1}{c}{mAP} & \multicolumn{1}{c}{R1} \\
    \Xhline{0.6pt}
    VDT~\cite{zhang2024view}      & 55.20                               & 64.10                  & 42.76                   & 48.12                  & 66.83                   & \textbf{82.50}         & 71.59                   & 82.14                  \\
    DTST~\cite{wang2024dynamic}   & 55.73                               & 64.42                  & 43.49                   & 50.53                  & 63.31                   & 80.00                  & 72.40                   & 78.57                  \\
    VIF~\cite{khalid2025bridging} & 57.46                               & 65.71                  & 44.55                   & 51.25                  & 66.98                   & \textbf{82.50}         & 74.19                   & 83.93                  \\
    SD-ReID~\cite{wang2026sd}     & 57.47                               & 65.06                  & 46.44                   & 53.12                  & 67.70                   & \textbf{82.50}         & 74.08                   & 81.25                  \\
    GSAlign~\cite{li2026gsalign}  & 57.95                               & 65.06                  & 61.55                   & 64.89                  & 65.55                   & 80.00                  & 73.86                   & 83.04                  \\
    SAS-VPReID~\cite{yang2026sas} & 58.48                               & 64.42                  & 57.36                   & 58.51                  & 63.41                   & 70.00                  & 76.90                   & 83.93                  \\
    SeCap~\cite{wang2025secap}    & 60.19                               & 68.59                  & 58.94                   & 69.43                  & 68.08                   & 80.00                  & 75.42                   & 86.61                  \\
    ViSA~\cite{zhang2026view}     & 65.46                               & 70.51                  & 69.00                   & 71.28                  & 67.78                   & \textbf{82.50}         & \textbf{83.90}          & 88.39                  \\
    LATex~\cite{zhang2025latex}   & 67.09                               & \textbf{76.96}         & 58.88                   & 66.87                  & 69.06                   & 80.00                  & 79.90                   & \textbf{90.18}         \\
    \Xhline{0.6pt}
    \rowcolor{gray!15}
    \textbf{HiHR}                 & \textbf{67.52}                      & 74.36                  & \textbf{70.53}          & \textbf{75.53}         & \textbf{69.60}          & 75.00                  & 80.93                   & 87.50                  \\
    \Xhline{1.0pt}
  \end{tabular}
  \vspace{-2mm}
\end{table}
\subsection{Ablation Study}
\label{sec:ablation}
In this section, we conduct experiments on CARGO to verify the effect of each design in our framework.
We employ the cross-entropy loss and triplet loss to train the baseline model, which is based on the ViT-B/16 from CLIP.

\begin{table}[t]
  \caption{\textbf{Ablation study of each module on CARGO.}
  }
  \label{tab:ablation_components}
  \centering
  \begin{tabular}{l|cc|cc|cc|cc}
    \Xhline{1.0pt}
    \multirow{2}{*}{Method}
                     & \multicolumn{2}{c|}{ALL}
                     & \multicolumn{2}{c|}{A$\rightarrow$G}
                     & \multicolumn{2}{c|}{A$\rightarrow$A}
                     & \multicolumn{2}{c}{G$\rightarrow$G}                                                                                                                         \\
    \cline{2-3}\cline{4-5}\cline{6-7}\cline{8-9}
                     & mAP                                  & R1             & mAP            & R1             & mAP            & R1             & mAP            & R1             \\
    \Xhline{0.6pt}
    Baseline         & 63.61                                & 68.23          & 66.06          & 66.02          & 64.70          & 71.50          & 79.08          & 87.39          \\
    Baseline+TMF     & 65.46                                & 72.08          & 69.50          & 70.28          & 65.64          & 71.50          & 80.22          & \textbf{88.29} \\
    \rowcolor{gray!15}
    Baseline+TMF+HHL & \textbf{67.52}                       & \textbf{74.36} & \textbf{70.53} & \textbf{75.53} & \textbf{69.60} & \textbf{75.00} & \textbf{80.93} & 87.50          \\
    \Xhline{1.0pt}
  \end{tabular}
  \vspace{-2mm}
\end{table}

\textbf{Component Analysis.}
As shown in Tab.~\ref{tab:ablation_components}, adding TMF brings clear performance gains over the baseline, especially under the challenging A$\rightarrow$G setting (+3.44 mAP and +4.26 Rank-1).
This improvement verifies the value of TMF in extracting more discriminative features.
Further incorporating HHL consistently improves mAP across protocols, showing that hierarchical supervision complements TMF by preserving view-specific discriminative cues while avoiding the cross-view feature collapse.
The above results fully demonstrate the effectiveness of our key modules.
\begin{figure}[tb]
  \centering
  \includegraphics[width=0.9\columnwidth]{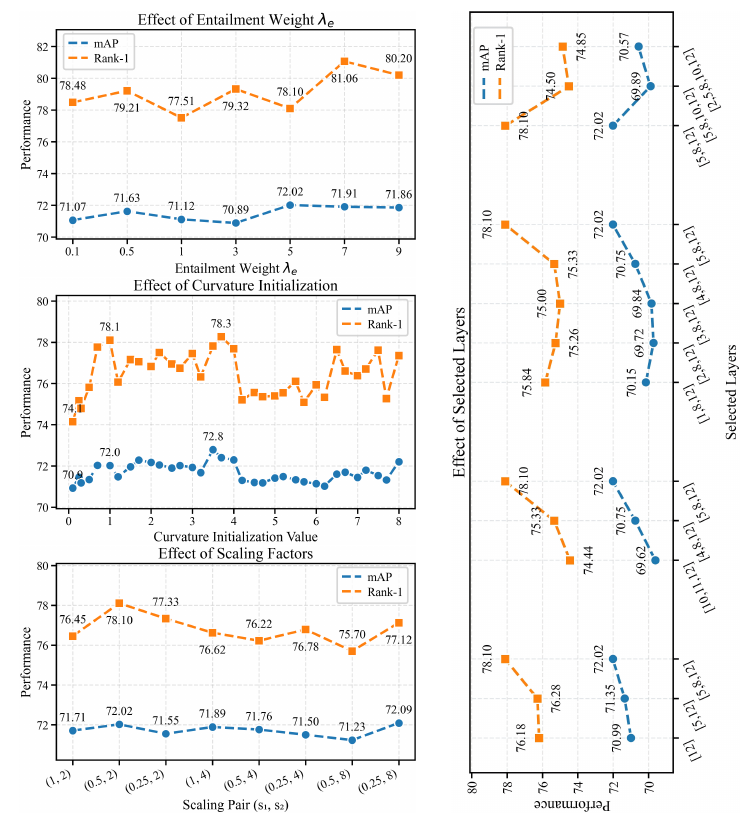}
  \caption{\textbf{Hyperparameter and layer-selection analysis on CARGO.}
    The left panels report the entailment weight $\lambda_e$, curvature initialization $\tau$, and scaling factors $(s_1,s_2)$ from top to bottom.
    The right panel reports selected Transformer layers.}
  \label{fig:ablation_of_curvature_and_scaler}
  \vspace{-4mm}
\end{figure}

\textbf{Hyperparameter Sensitivity Analysis.}
As shown in the left panels of Fig.~\ref{fig:ablation_of_curvature_and_scaler}, our method maintains stable performance across different entailment weight $\lambda_e$, curvature initialization $\tau$, and scaling factors $(s_1,s_2)$.
For the entailment weight $\lambda_e$, the performance remains consistently stable across a wide range of values.
As shown in Fig.~\ref{fig:ablation_of_curvature_and_scaler}, mAP fluctuates slightly around 71--72, and Rank-1 also stays within a narrow performance band.
This indicates that our method is insensitive to the choice of $\lambda_e$.
For the curvature $\tau$, initialization values from 1.0 to 4.0 consistently yield strong mAP and Rank-1, whereas small curvatures lead to a noticeable performance degradation.
This suggests that sufficient a large negative curvature is necessary to provide the capacity for hierarchical organization and child-cluster separation.
Once such geometry is properly initialized, the model can adapt the curvature during training and is not sensitive to its exact starting value.
For the scaling factors $(s_1,s_2)$, different initializations produce a similar final accuracy.
It indicates that the parent-child scale separation can be reliably learned from diverse starting points.
Overall, these results show that our method is robust to key hyper-parameters and does not require delicate tuning.

\textbf{Layer Selection Analysis.}
The right panel of Fig.~\ref{fig:ablation_of_curvature_and_scaler} evaluates the effect of selected Transformer layers on multi-granularity feature extraction.
Using only the final layer $\{12\}$ is suboptimal, while incorporating intermediate layers improves performance, with $\{5,8,12\}$ achieving the best results.
This shows that mid-level features provide complementary information to high-level features for AG-ReID.
However, the improvement depends on layer complementarity rather than the number of selected layers.
The deep-stacked setting $\{10,11,12\}$ performs worse than the cross-depth setting $\{5,8,12\}$, indicating that adjacent high-level layers contain redundant semantics and lack sufficient granularity diversity.
Shallow combinations are also less effective, since early layers mainly encode low-level patterns that are weakly aligned with identity-discriminative semantics.
Moreover, adding more layers beyond $\{5,8,12\}$ does not bring further gains and can even degrade performance, suggesting indiscriminate multi-layer fusion introduces redundancy and noisy interactions.
These results validate the compact cross-depth design of multi-granularity feature extraction, which captures complementary semantic and local cues.
\begin{table}[tb]
  \caption{\textbf{Ablation study of loss functions on CARGO.}
  }
  \label{tab:ablation_loss}
  \centering
  \resizebox{\columnwidth}{!}{
    \begin{tabular}{ccc|cc|cc|cc|cc}
      \Xhline{1.0pt}
      \multirow{2}{*}{$\mathcal{L}_{\mathrm{e}}(\mathbf{z}^{\boldsymbol{\pi}},\mathbf{z}^{c})$}
                 & \multirow{2}{*}{$\mathcal{L}_{\mathrm{e}}(\mathbf{z}^{p}_{P},\mathbf{z}^{c}_{P})$}
                 & \multirow{2}{*}{$\mathcal{L}_{\mathrm{e}}(\mathbf{z}^{p}_{P},\mathbf{z}^{p})+\mathcal{L}_{\mathrm{e}}(\mathbf{z}^{c}_{P},\mathbf{z}^{c})$}
                 & \multicolumn{2}{c|}{ALL}
                 & \multicolumn{2}{c|}{A$\rightarrow$G}
                 & \multicolumn{2}{c|}{A$\rightarrow$A}
                 & \multicolumn{2}{c}{G$\rightarrow$G}                                                                                                                                                                                                                                                             \\
      \cline{4-5}\cline{6-7}\cline{8-9}\cline{10-11}
                 &                                                                                                                                            &            & mAP            & R1             & mAP            & R1             & mAP            & R1             & mAP            & R1             \\
      \Xhline{0.6pt}
                 &                                                                                                                                            &            & 64.62          & 71.15          & 68.06          & 72.34          & 67.42          & \textbf{77.50} & 79.80          & 87.50          \\
      \checkmark &                                                                                                                                            &            & 65.46          & 72.12          & 69.50          & 70.21          & 65.64          & 72.50          & 80.22          & \textbf{88.39} \\
      \checkmark & \checkmark                                                                                                                                 &            & 65.71          & 72.44          & 68.21          & 71.28          & 67.49          & \textbf{77.50} & 80.65          & 87.50          \\
      \rowcolor{gray!15}
      \checkmark & \checkmark                                                                                                                                 & \checkmark & \textbf{67.52} & \textbf{74.36} & \textbf{70.53} & \textbf{75.53} & \textbf{69.60} & 75.00          & \textbf{80.93} & 87.50          \\
      \Xhline{1.0pt}
    \end{tabular}
  }
\end{table}

\textbf{Loss Function Analysis.}
As shown in Tab.~\ref{tab:ablation_loss}, the model without entailment losses achieves the lowest overall performance, indicating that the metric supervision alone is insufficient to maintain the hierarchical structure.
Introducing $\mathcal{L}_{\mathrm{e}}(\mathbf{z}^{\boldsymbol{\pi}},\mathbf{z}^{c})$ improves the ALL performance from 64.62/71.15 to 65.46/72.12, showing that constraining child embeddings within parent-induced cones helps reduce child-cluster drift.
Adding $\mathcal{L}_{\mathrm{e}}(\mathbf{z}^{p}_{P},\mathbf{z}^{c}_{P})$ further improves the results by enforcing consistency between parent and child prompts.
When the consistency terms $\mathcal{L}_{\mathrm{e}}(\mathbf{z}^{p}_{P},\mathbf{z}^{p})+\mathcal{L}_{\mathrm{e}}(\mathbf{z}^{c}_{P},\mathbf{z}^{c})$ are enabled, the model achieves the best ALL performance.
These results demonstrate that the four entailment constraints are complementary for hierarchical representation learning.
\begin{table}[t]
  \caption{\textbf{Ablation study of the prompt design on CARGO.}
  }
  \label{tab:ablation_prompt}
  \centering
  \begin{tabular}{cc|cc|cc|cc|cc}
    \Xhline{1.0pt}
    \multirow{2}{*}{Model}
      & \multirow{2}{*}{Prompt Type}
      & \multicolumn{2}{c|}{ALL}
      & \multicolumn{2}{c|}{A$\rightarrow$G}
      & \multicolumn{2}{c|}{A$\rightarrow$A}
      & \multicolumn{2}{c}{G$\rightarrow$G}                                                                                                                                          \\
    \cline{3-4}\cline{5-6}\cline{7-8}\cline{9-10}
      &                                      & mAP            & R1             & mAP            & R1             & mAP            & R1             & mAP            & R1             \\
    \Xhline{0.6pt}
    A & Text-only                            & 64.58          & 68.91          & 67.49          & 71.28          & 66.60          & 75.00          & 80.01          & 85.71          \\
    B & Text+CoCoOp                          & 66.73          & 71.47          & 68.19          & 72.34          & 68.01          & \textbf{77.50} & \textbf{81.22} & 85.71          \\
    \rowcolor{gray!15}
    C & Ours                                 & \textbf{67.52} & \textbf{74.36} & \textbf{70.53} & \textbf{75.53} & \textbf{69.60} & 75.00          & 80.93          & \textbf{87.50} \\
    \Xhline{1.0pt}
  \end{tabular}
\end{table}

\textbf{Prompt Design Analysis.}
Tab.~\ref{tab:ablation_prompt} compares three prompt designs.
Model A uses fixed handcrafted prompts, where $\mathbf{p}_{vag}$ is ``\textit{A photo of a person.}'' and $\mathbf{p}_{vaw}$ is ``\textit{A photo of a person from [an aerial/a ground] view.}''.
Model B adopts CoCoOp~\cite{zhou2022conditional} to generate instance-conditional prompts, while Model C uses our dual-level prompts.
As shown in Tab.~\ref{tab:ablation_prompt}, Model A obtains reasonable performance, indicating that explicit textual semantics benefit multi-granularity feature fusion.
Model B improves over Model A, showing the effect of learnable prompt adaptation.
Compared with Model B, Model C achieves better results, improving ALL performance to 67.52\% mAP and 74.36\% Rank-1.
This suggests that compact dataset-level prompts are more effective for AG-ReID than instance-wise prompt generation.
The reason is that aerial-ground discrepancies are mainly governed by systematic view-related variations rather than highly sample-specific changes.
\begin{table}[!t]
  \caption{\textbf{Ablation study of the fusion design on CARGO.}
  }
  \label{tab:ablation_fusion}
  \centering
  \begin{tabular}{cc|cc|cc|cc|cc}
    \Xhline{1.0pt}
    \multirow{2}{*}{Model}
      & \multirow{2}{*}{Fusion Type}
      & \multicolumn{2}{c|}{ALL}
      & \multicolumn{2}{c|}{A$\rightarrow$G}
      & \multicolumn{2}{c|}{A$\rightarrow$A}
      & \multicolumn{2}{c}{G$\rightarrow$G}                                                                                                                                          \\
    \cline{3-4}\cline{5-6}\cline{7-8}\cline{9-10}
      &                                      & mAP            & R1             & mAP            & R1             & mAP            & R1             & mAP            & R1             \\
    \Xhline{0.6pt}
    A & Mean                                 & 64.95          & 69.87          & 66.86          & 69.15          & 65.49          & 72.50          & 80.16          & 87.50          \\
    B & CA                                   & 65.80          & 71.15          & 67.08          & 70.21          & 67.20          & \textbf{77.50} & 80.75          & 86.61          \\
    \rowcolor{gray!15}
    C & TMF                                  & \textbf{67.52} & \textbf{74.36} & \textbf{70.53} & \textbf{75.53} & \textbf{69.60} & 75.00          & \textbf{80.93} & \textbf{87.50} \\
    \Xhline{1.0pt}
  \end{tabular}
\end{table}

\textbf{Fusion Design Analysis.}
Tab.~\ref{tab:ablation_fusion} compares three fusion designs.
Model A uses a mean aggregation over class tokens from selected layers.
Model B applies a standard cross-attention between prompts and class tokens.
Model C denotes our proposed TMF.
As shown in Tab.~\ref{tab:ablation_fusion}, Model A already achieves reasonable performance, confirming the usefulness of multi-layer visual cues.
Model B improves over Model A, indicating that an adaptive prompt-token interaction is more effective than static averaging.
Our TMF obtains the best results, outperforming Model B by +1.72 mAP and +3.21 Rank-1 under the ALL setting.
This verifies that the proposed TMF more effectively exploits complementary information across layers for AG-ReID.
\section{Conclusion}
\label{sec:conclusion}
In this paper, we propose HiHR, a hierarchical hyperbolic representation framework for jointly exploiting view-invariant identity semantics and view-specific discriminative details.
Specifically, we first extract visual-text features at multiple granularities.
Then, we propose TMF to fuse these multi-granularity features, thereby enhancing the representation ability of identity features.
Furthermore, we introduce HHL to construct a hierarchical feature structure within the hyperbolic space.
The resulting hierarchy consists of a coarse level for ensuring identity separability and cross-view consistency, and a fine level for preserving view-specific discriminative cues.
As a result, our proposed framework can effectively aggregate view-invariant and view-specific discriminative features for AG-ReID.
Extensive experiments on four AG-ReID benchmarks demonstrate that our framework achieves competitive or state-of-the-art performance.
%

\bibliographystyle{splncs04}
\bibliography{main}

\end{document}